\documentclass[letterpaper,10pt,conference]{ieeeconf}
\IEEEoverridecommandlockouts
\usepackage[T1]{fontenc}
\usepackage{float}
\usepackage{amsfonts}
\usepackage{amsthm}
\usepackage{pifont}
\usepackage{xcolor}
\usepackage{soul}

\usepackage{glossaries}
\usepackage{listings}
\usepackage{amsmath}
\usepackage{amssymb}
\usepackage{subfig}
\usepackage{graphicx}
\usepackage{stix}
\usepackage{tikz}
\usepackage{etoolbox}
\usetikzlibrary{shapes,arrows}
\usepackage{verbatim}
\usepackage[hidelinks]{hyperref}

\renewcommand{\baselinestretch}{1.04}

\newcommand{\deleted}[1]{}

\begin{document}
\title{\bf 
PogoDrone: Design, Model, and Control  of a Jumping Quadrotor
}
\author{Brian Zhu, Jiawei Xu, Andrew Charway, and David Saldaña
\thanks{
B. Zhu, J. Xu , and D. Salda\~{n}a are with the Autonomous and Intelligent Robotics Laboratory (AIRLab), Lehigh University, PA, USA. Email:~$\{$\texttt{blz222, jix519, anc619, saldana\}@lehigh.edu}}
}
\maketitle
\begin{abstract}
We present a design, model, and control for a novel jumping-flying robot that is called PogoDrone. The robot is composed of a quadrotor with a passive mechanism for jumping.
The robot can continuously jump in place or fly like a normal quadrotor. Jumping in place allows the robot to quickly move and operate very close to the ground.
For instance, in agricultural applications, the jumping mechanism allows the robot to take samples of soil.
We propose a hybrid controller that switches from attitude to position control to allow the robot to fall horizontally and recover to the original position.
We compare the jumping mode with the hovering mode to analyze the energy consumption.
In simulations, we evaluate the effect of different factors
on energy consumption. In real experiments, we show that our robot can repeatedly impact the ground, jump, and fly in a physical environment.
%

\end{abstract}

\section{Introduction}
\noindent
In robotics, jumping mechanisms have been introduced based on bio-inspired locomotion principles~\cite{article,Truong_2019}. 
A jumping robot has a strong ability to overcome high obstacles~\cite{article}, but it is unable to stay in the air. 
Although many researchers have investigated miniature jumping robots over the last decade~\cite{6204349, woodward2011design, kovac2008miniature}, only a few have shown the integration of other locomotion modes. The  jumpglider~\cite{6181502} is one of the hybrid robots that successfully achieved the integration of two locomotion modes. Their robot is equipped with foldable gliding wings that can improve travel distance and reduce the impact on landing. Though jumping robots are able to move in a complex changeable environment with high obstacles, the height of their jumps is limited by their structural design.
For example, the flea-inspired jumping robot, designed by Noh et al.~\cite{6204349}, can jump over a height of no more than 30 times its body height.

Quadrotors have become popular across various industries and as a research topic in recent years. Some of the common applications include search and rescue operations~\cite{naidoo2011development}, exploration, aerial surveillance~\cite{LAGRING2012644}, and transportation~\cite{loianno2017cooperative}. One challenge that quadrotors must face is the ground effect~\cite{powers2013influence}, which inhibits the vehicle from operating close to the ground due to the turbulent flow produced by its rotors.
Additionally, a major challenge that researchers have to overcome is the short flight time~\cite{BOUKOBERINE2019113823}, which is typically around 5-30 minutes. To deal with this challenge, researchers have come up with strategies and techniques including using power sources with a higher energy density such as fuel cells~\cite{10.1115/IMECE2012-88871}, applying laser-beam in-flight recharging~\cite{8403572,6094731}, connecting with a power source through long-range tethering~\cite{MUTTIN2011332}, and swapping batteries in flight~\cite{9197580}.

Although adding a gliding capability to a jumping robot \cite{6181502} brings together the benefits of fast travel over a long distance in-air and energy-efficient obstacle clearance, the vehicle is not able to actuate in the air.
The integration of multi-rotors in a jumping system gives it more maneuverability in flight. 
In a recent publication~\cite{haldane2017repetitive}, the authors propose an active jumping mechanism with two rotors and a tail. The rotors are used to control roll and yaw angles, but they do not help the robot to stay in the air.

%
\begin{figure}[t!]
    \includegraphics[width=8.6cm]{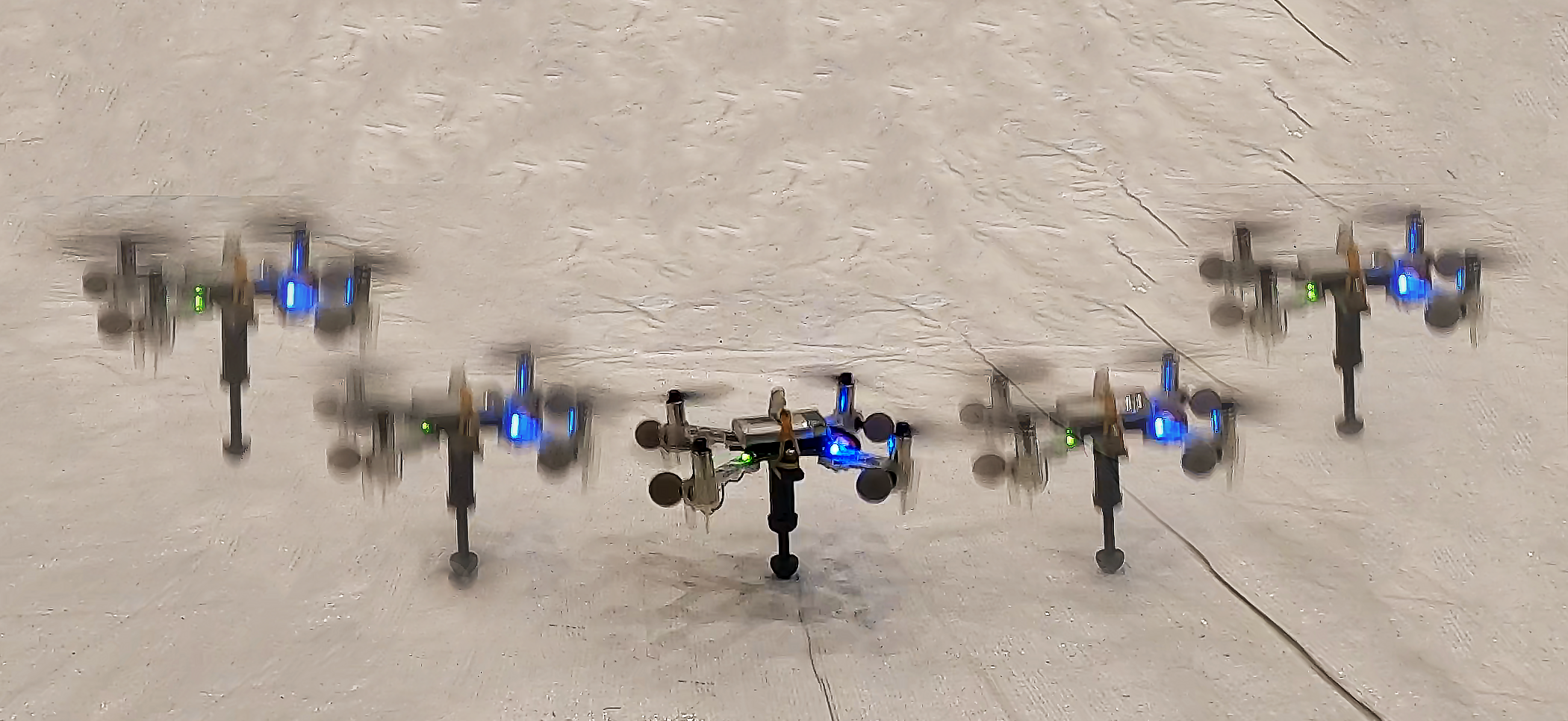}
    \caption{A PogoDrone during a bounce. It descends, compresses, rebounds, and flies back to its original location. The PogoDrone at the center shows the case when the spring is fully compressed.}
    \vspace{-1em}
    \label{fig:titlePic}
\end{figure}

Inspired by the combination of jumping and aerial robots, we propose a jumping-flying robot, called the \textit{PogoDrone}. 
We design, model, and control a novel robot that combines a quadrotor with a spring-based mechanism.
Fig.~\ref{fig:titlePic} shows the PogoDrone during a jump.
One of the main principles of creating this hybrid system is the conservation of energy. Typically, when a normal quadrotor drops from a certain height and collides with the ground, the kinematic energy is dissipated through the collision. In contrast, when a PogoDrone makes contact with the ground, the kinetic energy is stored through the compression of a spring as potential energy. Upon decompression, the spring releases the energy back to the PogoDrone. Note that not all the energy is conserved in such a process because of the imperfection of the spring. Thus, we still need to compensate for the loss of energy by actuating the rotors.
A PogoDrone requires less power to function due to the passive jumping capability that preserves energy, which increases energy efficiency and the maximum flight time.

In comparison to a single-mode jumping robot, such as the Salto-1P \cite{haldane2017repetitive}, we offer a significantly simpler design that can jump and fly continuously. The simplicity gives our robot superior maneuverability which actively flying and addressing the Salto-1P's inability to perform in scenarios that require continuous hovering. Compared to a traditional quadrotor, our PogoDrone allows operations close to the ground by deactivating its rotors when descending, which mitigates the ground effect problem~\cite{kushleyev2013towards}.







\section{Design}
\label{sec:Design}
\noindent 
Our robot design adopts one of the simplest and most common methods to store potential energy, a spring in a pogo-stick.
A PogoDrone is composed of a quadrotor fitted with a miniature and lightweight pogo-stick. The main components are described as follows.
\paragraph{Flying Vehicle} 
The quadrotor platform used for the PogoDrone is the Crazyflie 2.1. Open-source software and hardware, and its popularity in aerial robotics research make it an ideal choice for this project. The vehicle weighs 27g with the battery, and can carry a maximum payload of 15g. The dimensions are $92\times92\times29$~mm, and the 1-cell LiPo battery allows up to five minutes of hover time in the air. 


\begin{figure}[t]
\centering
    \includegraphics[width=0.8\linewidth]{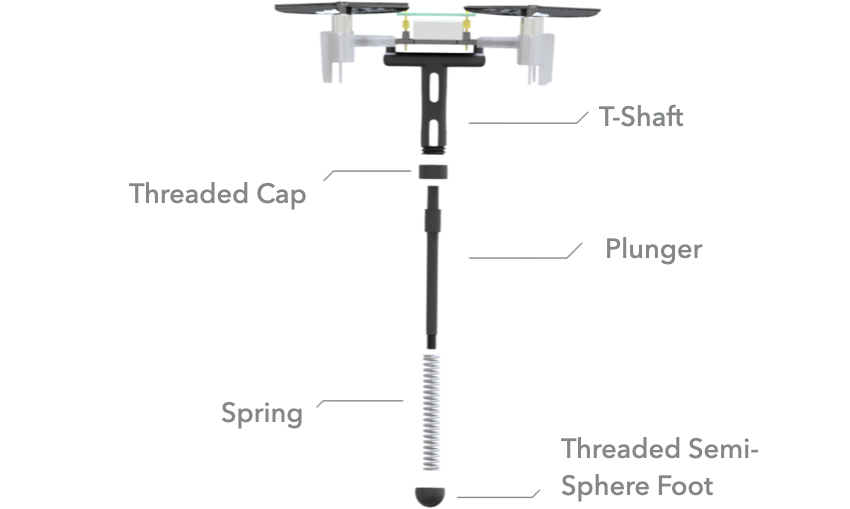}
    \caption{Components of the PogoDrone}
    \vspace{-1em}
    \label{fig:pogodroneCAD}
\end{figure}
\paragraph{Pogo-stick Mechanism}
We show the parts of the miniature pogo-stick in Fig.~\ref{fig:pogodroneCAD}. 
The dimensions of the pogo-stick are $35.25\times11\times55$mm. It is comprised of four 3-D printed parts along with a spring.
The spring's length is $17.88$mm. 
The spring constant was obtained experimentally to be $394.58$N/m using weights and calipers. Using built-in mounting holes and a lightweight skeletonized frame, the primary T-shaped housing of the pogo-stick extends down the center of mass by approximately the same distance as the spring length, keeping the spring from generating unwanted torque when compressed. Our spring length, however, is constrained by the instability of the Crazyflie quadrotor. The longer the spring length, the longer the shaft and the lower the center of gravity. This increases the inertia tensor of the system which causes small changes in angular acceleration to generate large magnitudes of torque, thus making the system extremely difficult to fly. 
A plunger, held in by a threaded cap, inserts into the main shaft, compressing the spring on impact. Lastly, the semi-sphere foot at the end of the pogo-stick allows rotational motion from a pivot point. Given tight weight constraints, lightweight design of the pogo-stick is integral. Our current design weighs 4g, which is under the payload capacity of the Crazyflie (15g).


\section{Model}
\label{sec:Model}
\begin{figure}[t!]
    \center
    \includegraphics[width=.7\linewidth]{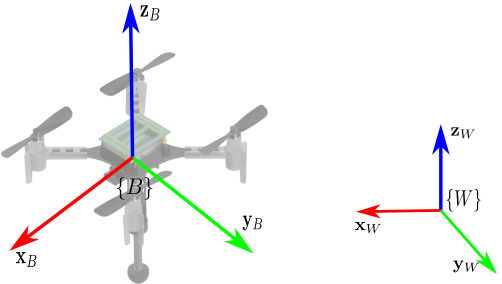}
    \caption{The PogoDrone with its coordinate frame}
    \vspace{-1em}
    \label{fig:coordinatePic}
\end{figure}
\noindent
Our PogoDrone is composed of a quadrotor with a bottom side attached jumping device, modeled as a spring-damper. 
The robot has a mass $m$ and an inertia tensor $\boldsymbol I$. Since the mass of the jumping mechanism is small in comparison with the quadrotor, we assume that the center of mass of the PogoDrone is at the same location as the center of mass of the quadrotor.
The coordinate frame for the quadrotor is denoted by $\{B\}$ and its origin is at the center of mass of the quadrotor. 
The $x$-axis is aligned with the front of the quadrotor and the $z$-axis is pointing upwards. 
The spring-damper mechanism has one of its ends attached to the origin of $\{B\}$ and extends along the negative $z$-axis of $\{B\}$ (see Fig.~\ref{fig:coordinatePic}). The pogo has a natural length $l_0$ and a minimum length $l_{min}$.
The world coordinate frame is denoted by $\{W\}$ with the $z$-axis pointing upwards. We denote the rotation matrix from body frame to world frame as~${}^W\!\!\boldsymbol{R}\!_B\in\mathsf{SO(3)}$.

\subsection{Sensors and Actuators}
\noindent
We use internal and external sensors to measure the location of the robot $\boldsymbol{r}\in \mathbb{R}^3$ in $\{W\}$, velocity $\boldsymbol{v}\in\mathbb{R}^3$, attitude ${}^W\!\!\boldsymbol{R}\!_B$, and angular velocity $\boldsymbol{\omega}\in\mathbb{R}^3$.
The robot has four motor, as active actuators, that generate a thrust $f_q\in \mathbb{R}$ along the $z$-axis of the body frame and a torque $\boldsymbol\tau \in \mathbb{R}^3$.
Therefore, our control input is the tuple $(f_q, \boldsymbol\tau)$ which, through motor power distribution, can map to individual motor forces~\cite{5717652}.
In addition, the pogo mechanism, as a passive actuator, generates a force,
\begin{equation}
f_{s} = - k\Delta L - b\frac{d}{dt}\Delta L,
\label{eq:spring}
\end{equation}
where $k>0$ is the spring constant, $b>0$ is the damping factor, and $\Delta L=l - l_0$ is the amount of deformation of the spring. Since the spring is aligned with $z$-axis of the quadrotor, it does not generate any torque. Although the robot design has a spring and no damper, we included the damping term to take into account the energy lost due to friction and spring imperfections.
The total force of the robot is along $z$-axis in~$\{B\}$,
\begin{equation}
f = f_q + f_s.
\end{equation}

\subsection{Dynamics}
\noindent
The dynamics of the robot depends on whether or not the PogoDrone makes a contact with the floor. The robot can be either in \textit{a)} a flying state or \textit{b)} a contact state.
We describe the dynamics of the vehicle with Newton-Euler's equations.

\textit{a) Flying State}
\begin{eqnarray}
    m\,\boldsymbol{\Ddot{r}}+m\,g\boldsymbol{  e}_3 &=& {f}\: {}^W\!\!\boldsymbol{R}\!_B\,\boldsymbol e_3\label{eq:newton},\\
    \boldsymbol{I}\Dot{\boldsymbol\omega}+\boldsymbol\omega\times \boldsymbol{I}\boldsymbol\omega &=& \boldsymbol\tau\label{eq:euler},
\label{eq:dynamics}
\end{eqnarray}
where $\boldsymbol{e}_3 = [0, 0, 1]^\top$ is a standard unit vector in $\mathbb{R}^3$ along $z$-axis. In this state, the robot does not have contact with the ground, and therefore, there is no spring compression, i.e,~$f_s = 0$.

\textit{b) Contact State}
During contact, the connection between the pogo and the floor can be modeled as a spherical joint. Therefore, the robot rotates around the contact point instead of the center of mass. The Newton equation \eqref{eq:newton} holds the same but we need to consider that force provided by the floor is transferred through the spring to the PogoDrone as $f_s\neq 0$. 
The Euler equation has some differences from \eqref{eq:euler} since the rotation point changed,
\begin{equation}
    \boldsymbol{I'\Dot\omega}+\boldsymbol\omega\times \boldsymbol{I'}\boldsymbol\omega = \boldsymbol\tau + {}^W\!\!\boldsymbol{R}_B(l_0+\Delta L)\boldsymbol{e}_3\times(-mg\boldsymbol{e}_3),
    \label{eq:neweuler}
\end{equation}
where $\boldsymbol{I'}$ is the moment of inertia about the contact point using the parallel axis theorem. In the experiments, the inertia is not adjusted online because the contact state is ephemeral, i.e., according to Section \ref{sec:Experiments}, the contact state lasts less than $100$ ms for each bounce. Therefore, we rely on the attitude controller to compensate for any shifts in the inertia matrix.
Note that there is an additional term due to the moment generated by the gravity force. 
We assume no slipping during the contact between the pogo tip and the ground.

\section{Control}
\label{sec:Control}
\begin{figure}[t]
    \includegraphics[width=\linewidth]{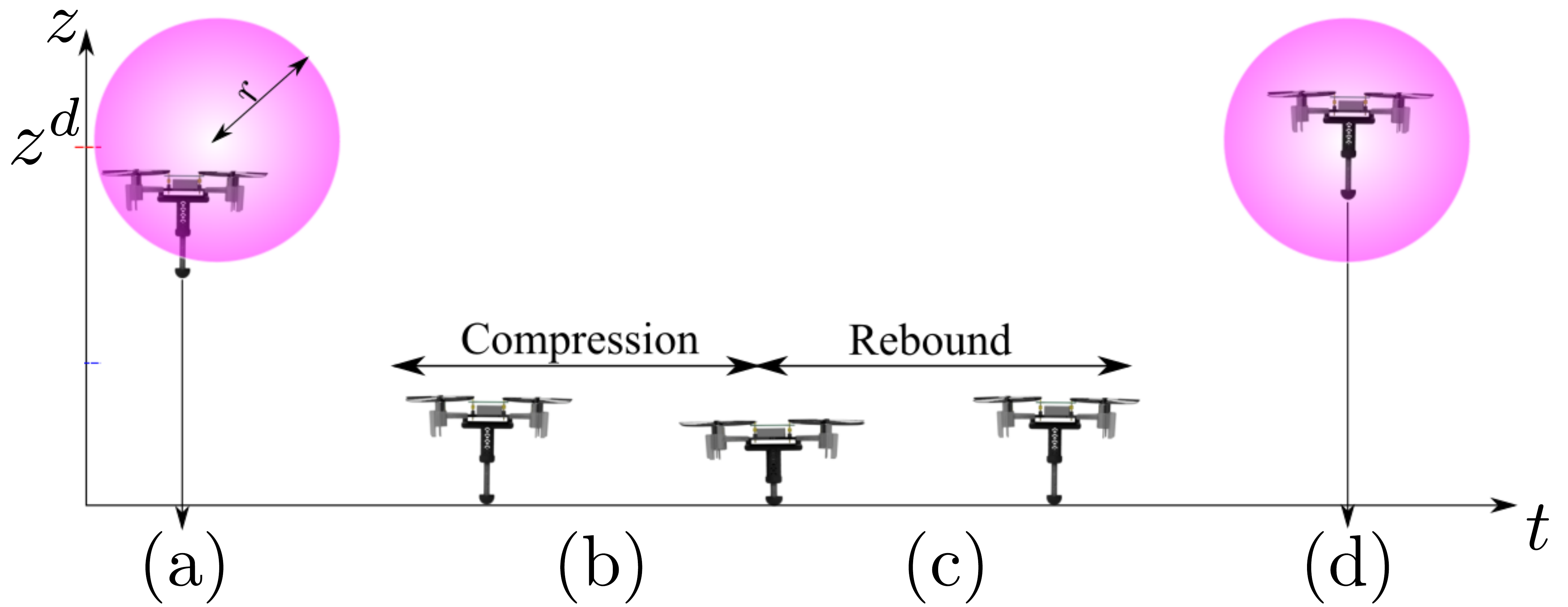}
    \caption{Phases of a jump: (a) descend, (b) compression, (c) rebound, and (d) ascend.}
    \label{fig:ControlPhases}
\end{figure}

\noindent
The control objective focuses on allowing the PogoDrone to jump repetitively from a desired position $\boldsymbol{r}^d$ with a desired yaw orientation $\psi^d$. Our robot passes through four phases during each jump (illustrated in Fig.~\ref{fig:ControlPhases}):\\
\textit{a) Descend: } The robot falls maintaining its attitude to approach the ground vertically, so the spring will be perpendicular to the ground. During this phase, the propellers are used to maintain attitude only, so the robot can fall vertically to convert as much gravitational potential energy into kinetic energy.\\
\textit{b) Compression: } The robot contacts with the ground and the spring compresses until it reaches its maximum compression length. During this phase, the spring converts the kinetic energy accumulated during the descend phase into elastic potential energy.\\
\textit{c) Rebound: } The spring expands, pushing the robot upwards. The spring converts the stored energy into kinetic energy. Note that we apply a realistic spring model in~\eqref{eq:spring}, meaning that there exists a loss of energy during the compression and rebound. The amount of energy that we can store with this strategy will depend on the length, stiffness of the spring, loss of energy due to friction, and spring imperfections.\\
\textit{d) Ascend: } The robot ascends to its original point $\boldsymbol{r}^d$. Since there is a loss in energy, the robot uses its propellers to reach the goal location.

\begin{figure}[t]
    \centering
    {\includegraphics[width=1.0\linewidth]{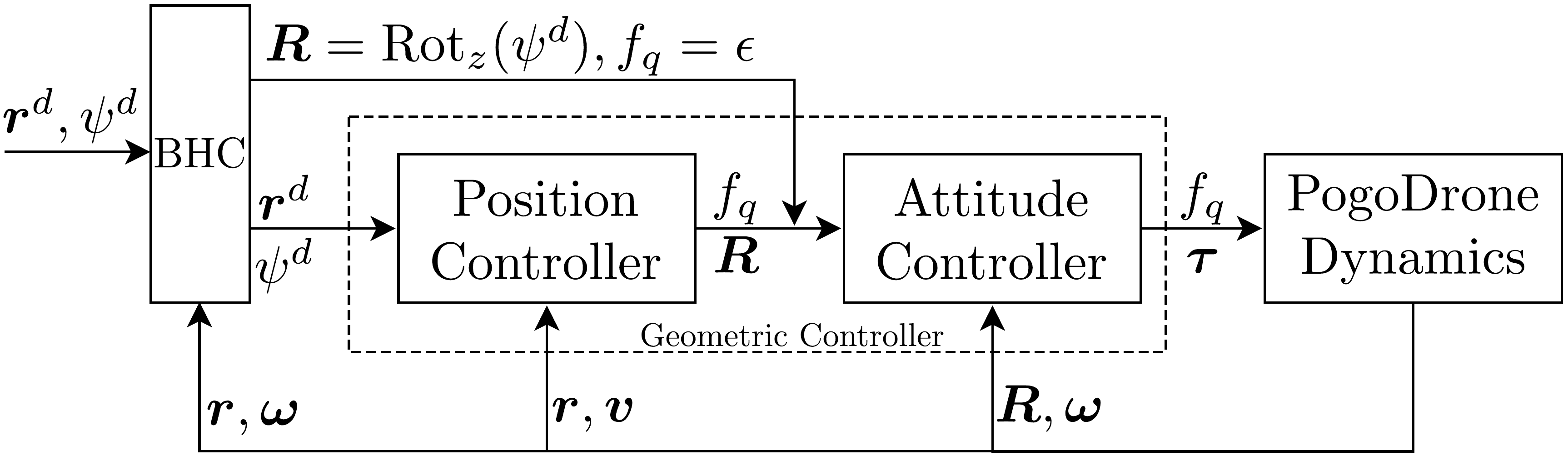}
    \caption{Control Diagram for the PogoDrone. The BHC block reads the current position and angular velocity of the PogoDrone and switches the control mode between the full geometric control and attitude-only control as described in Section \ref{sec:Control}. 
    }
    \vspace{-1em}
    \label{fig:Control}}
\end{figure}
\begin{figure}[b]
\centering
    \includegraphics[width=0.6\linewidth]{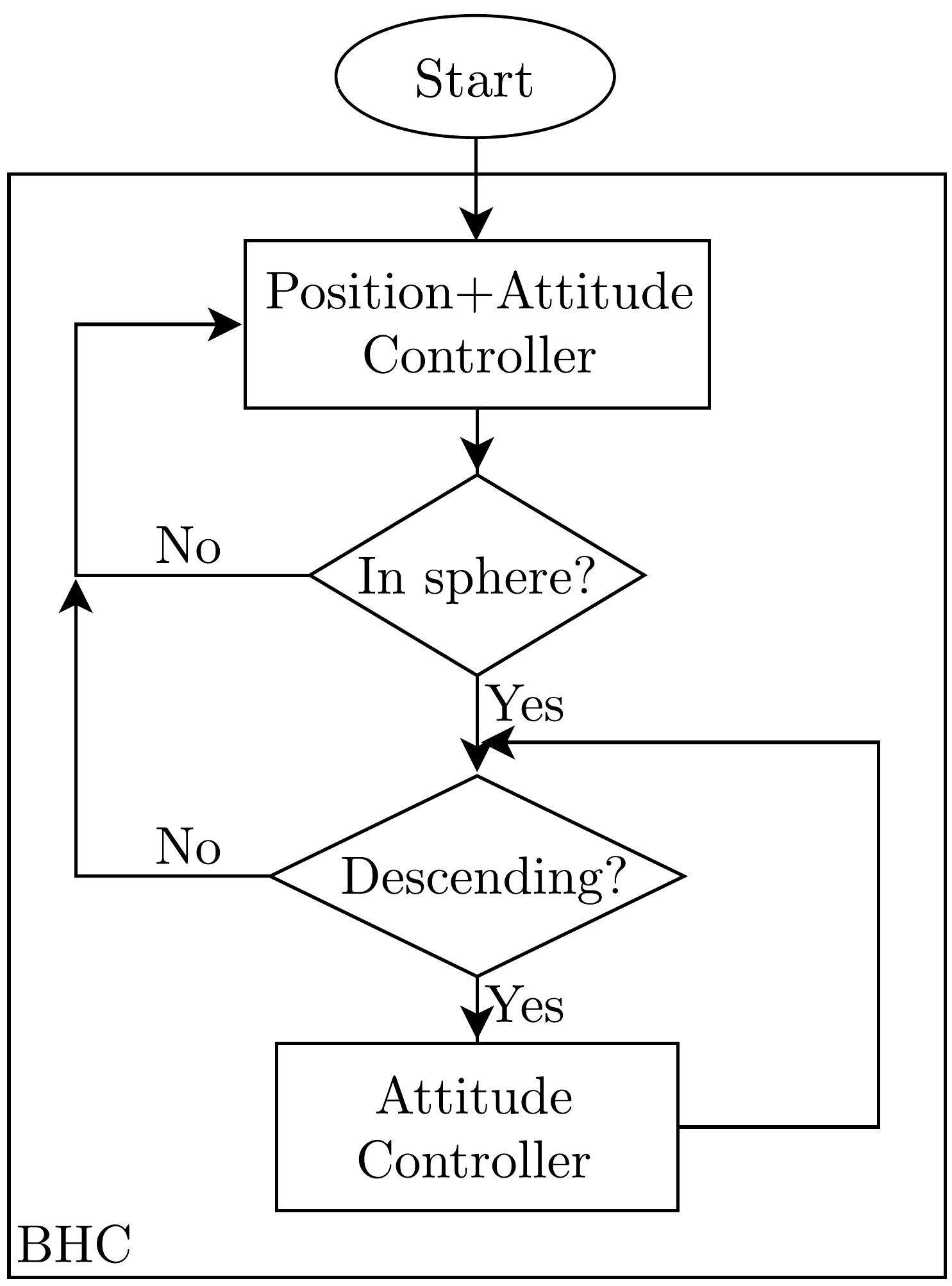}
    \caption{Flowchart of the Bounce-Hover Controller (BHC).}
    \label{fig:flowchart}
\end{figure}
We rely on the geometric  controller on $\mathsf{SE(3)}$~\cite{5717652, 5980409} for position and attitude control. We illustrate the control architecture on Fig.~\ref{fig:Control}.
Our module Bounce-Hover Controller (BHC) switches between the position and attitude controller depending on the phase of the jump. Our PogoDrone acts as a mass-damper-spring system when passively jumping, and as a quadrotor when actively hovering. 
During (a) descend and (b) compression, the robot does not need to compensate gravity or maintaining the position, so we use an attitude controller with $\boldsymbol{R}=\text{Rot}_z(\psi^d)$ to guarantee that the robot will stay horizontal, roll and pitch equal to zero, while holding its desired orientation in yaw.
Since we are not compensating for gravity, we do not need to generate a thrust force.
However, the geometric controller does not accept $f_q=0$ as an input, so we define an infinitesimally small constant  $\epsilon>0$ such that $f_q=\epsilon$.
During (c) rebound and (b) ascending, the PogoDrone uses the stored energy to push upwards 
and the quadrotor's force to fly back to the original position. For this purpose, the PogoDrone uses position control with $\boldsymbol{r}^d$ and $\psi^d$ as an input.

Our switching policy for the PogoDrone is defined by the algorithm in the BHC module.
We summarize our algorithm in the flowchart in Fig.~\ref{fig:flowchart}. 
The algorithm starts using the position controller to move the robot from any position to the desired position $\boldsymbol{r}^d$ with the desired orientation $\psi^d$.
Due to the error in the position controller, we use a sphere with center at center~$\boldsymbol{r}^d$, and radius $r>0$.
If the robot is within the sphere $\|\boldsymbol{r}^d-\boldsymbol{r}\|\leq r$ and the magnitude of its  angular velocity  $\|\boldsymbol{\omega}\|$ is less than a threshold, then it will start descending, leaving the sphere.
During the descend and compression, the robot uses the attitude controller until the rebound starts. At that moment, the robot switches from attitude to position control and comes back to the original position $\boldsymbol{r}^d$. This iterative process continues repeating periodically.

\section{Simulation and Experiments}
\noindent

We evaluate our PogoDrone model and design in simulations and actual robots\footnote{
        The simulator and ros packages can be found at: \url{https://github.com/swarmslab/PogoDrone}
    }.
We analyze the energy consumption  in a time interval $[t_0, t_f]$.
Since the current is proportional to the force generated by the spinning propellers, we use the rotors' force as a performance metric,
\begin{eqnarray}
e(t_f)= \int_{0}^{t_f} \sum_{i=1}^4 f_i(t) \: dt.
\label{eq:energy}
\end{eqnarray}
where $f_i$ is the force generated by the rotor $i$ at time $t$.
The final time $t_f$ defines the duration of the experiment with multiple jumps.

\subsection{Simulations}{
\label{sec:simulation}
\noindent
We compare the PogoDrone in jumping mode versus hovering mode. We run the simulations in batch to evaluate the factors that affect the energy consumption of the PogoDrone: \textit{a)} Noise level, \textit{b)} Spring constant, \textit{c)} Damping factor, \textit{d)} Hover height.

We developed a 2D simulation environment in Python. We implement the dynamics of the PogoDrone and its interaction with the ground surface as in Section \ref{sec:Model} using the Euler integration method with a time step of 1 millisecond. In the planar case, the PogoDrone has two propellers, making it an under-actuated system in 2-D, similar to a quadrotor in a 3-D environment. We first find a setting with the four factors where the PogoDrone
saves
energy with bouncing. Fig.~\ref{fig:simgood} shows the 2-norm of rotor forces, the position of the PogoDrone during the bounces, the spring lengths, and its forces. The numerical values of the four factors used in the reference is shown in the caption. Note that the values stated in Fig. \ref{fig:simgood} are close to, but not precisely the values used in the real robot experiments. The spring constant and hover height are comparable. It is challenging, however, to choose a spring with a specific damping factor since it is difficult to measure. Through different simulations and real experiments, we approximated a damping factor into the simulation that closely reflected the behavior of our real spring. For each numerical value selection of a factor, we run the simulation $20$ times for both normal hovering and bouncing mode while keeping all other factors invariant, and compare \eqref{eq:energy} from the two modes with $t_f = 18 s$.
\begin{figure}[t!]
\centering
    \includegraphics[width=\linewidth]{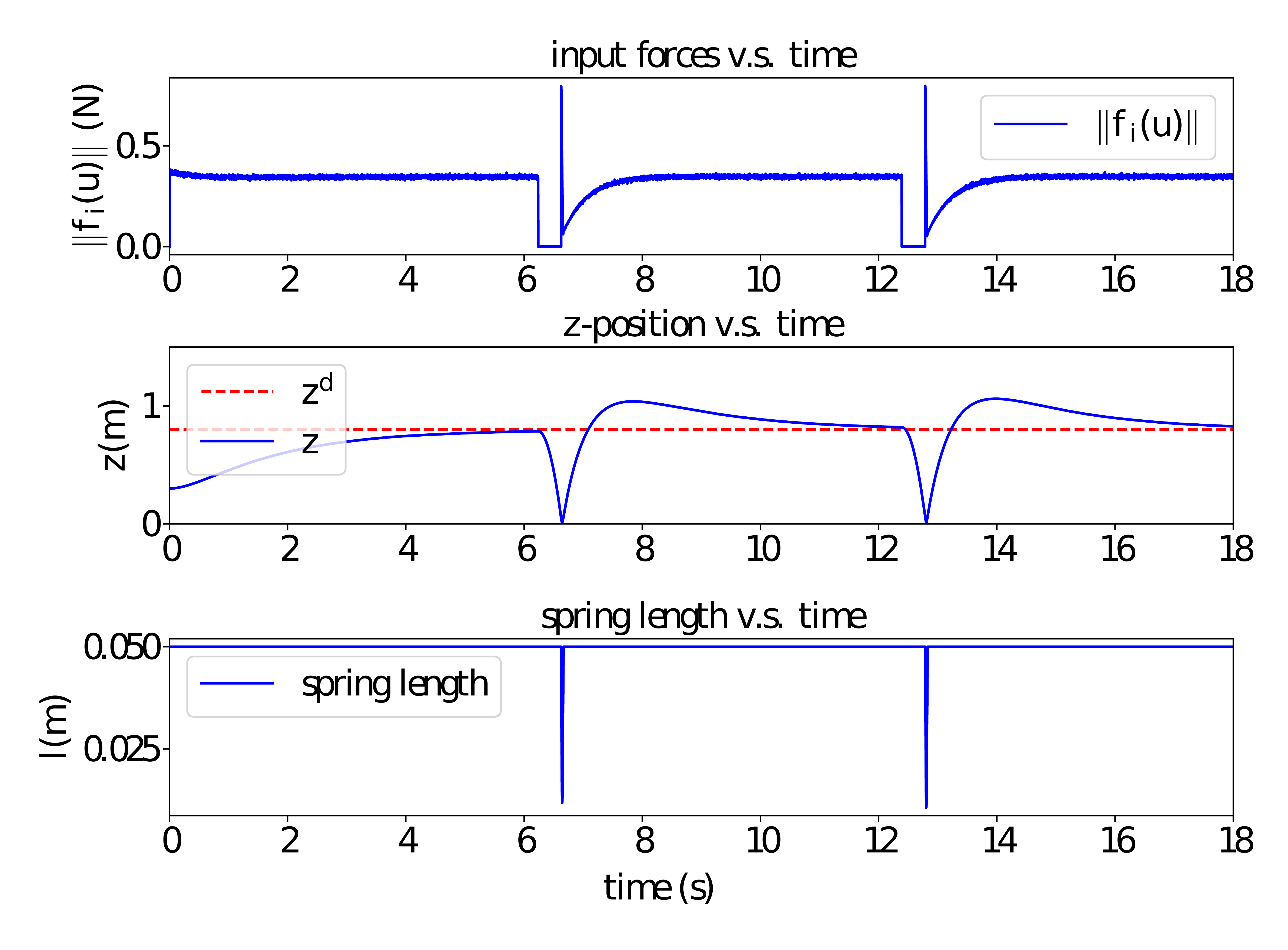}
    \caption{The performance of the reference simulation, which saves 12\% of the energy from hovering with two bounces. The numerical values of the four factors are: noise level = 0.2, spring constant = 400 N/m, damping factor = 1.0 Ns/m, hover height = 0.8 m.} 
    \label{fig:simgood}
\end{figure}

\noindent
\paragraph{Noise level}{
In the experiments with real robots, we notice two unmodelled variants in Section \ref{sec:Model} that keep the PogoDrone from achieving stability quickly. First, the uneven surface of the ground leads to the force received by the pogo having a perpendicular component to the pogo stick when touching the ground, causing large overshoots in $xy$-plane. Second, the force and torque generated by the propellers deviate from the desired input value $f_q$ and~$\boldsymbol{\tau}$ because of the imperfect motors, adding up to the loss of energy in achieving stability. As a result, the ascend phase lasts longer which keeps the motors actuating and increases the energy consumption. To emulate the errors from the input, we add Gaussian noise with standard deviation $\sigma$ to the propeller forces $f_i(\boldsymbol{u})$. For the imperfect ground, we add a Gaussian noise with standard deviation $5\sigma$ to the direction of the spring force such that the spring force aligns with $\text{Rot}_z(G){}^W\!\!\boldsymbol{R}\!_B\boldsymbol{e}_3$, where $G \sim \mathcal{N}(0, (5\sigma)^2)$, instead of ${}^W\!\!\boldsymbol{R}\!_B\boldsymbol{e}_3$.
We show in Fig. \ref{fig:simnl} that as the noise level increases, the energy consumption of both modes increases, and for the bouncing mode, it increases faster until catching up with the hovering mode. When the noise level is sufficiently high, the PogoDrone never achieves stability, making the propellers always actuated.
\begin{figure}[t!]
    \includegraphics[width=8.6cm]{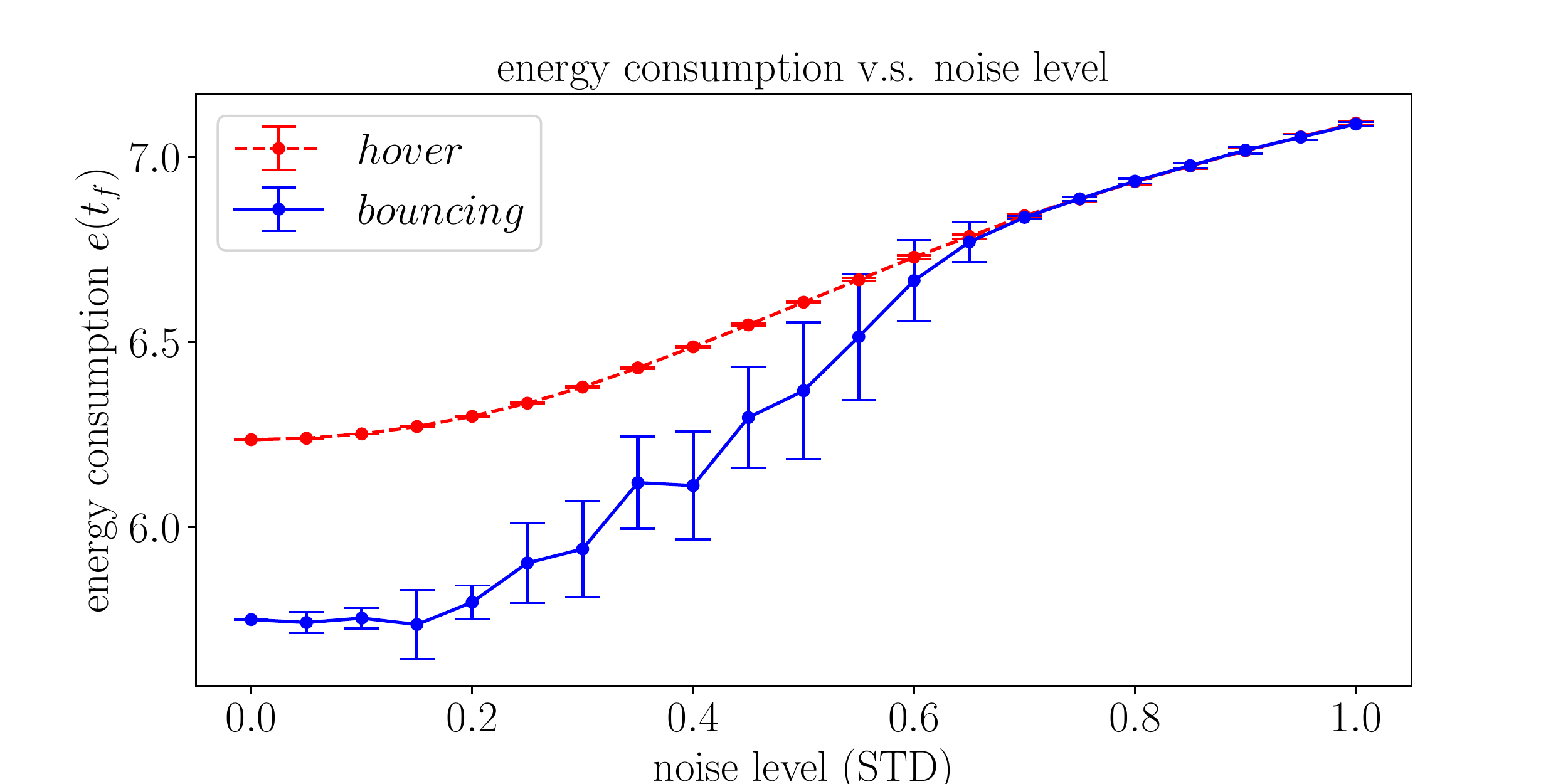}
    \caption{Energy consumption v.s. noise level. A Gaussian noise is applied on the thrust forces of the propellers and the force direction of the pogo, emulating imperfect motors and ground surface.}
    \label{fig:simnl}
\end{figure}
}

\noindent
\paragraph{Spring constant}{
The spring constant  determines the stiffness of the spring.
%
In Fig. \ref{fig:simsc}, we show the pattern of how the energy consumption changes as the spring constant gets higher. When the spring constant is low, the energy consumption decreases as the spring constant goes up because the spring can store an increasing amount of energy that would otherwise be lost during the hard impact. When the spring constant goes higher, the controller over-compensates the position in $z$, as shown in Fig. \ref{fig:simgood}. 
From 200 to 300 N/m is a critical interval (same as 700 to 900 {N/m}) because the number of jumps in the time interval can change depending on the noise. 
Note that between 300 and 700 N/m the energy consumption remains stable because the bouncing behavior couples with the controller tuning. After 900, the energy consumption slowly increases due to the damping term that prevents the spring from rebounding faster. When the spring constant is sufficiently high, then energy spent on recovering stability surpasses the energy conserved by the spring, which supports our hypothesis. The energy consumption of hovering is not affected because the pogo does not make contact with the ground.
\begin{figure}[t!]
    \includegraphics[width=8.6cm]{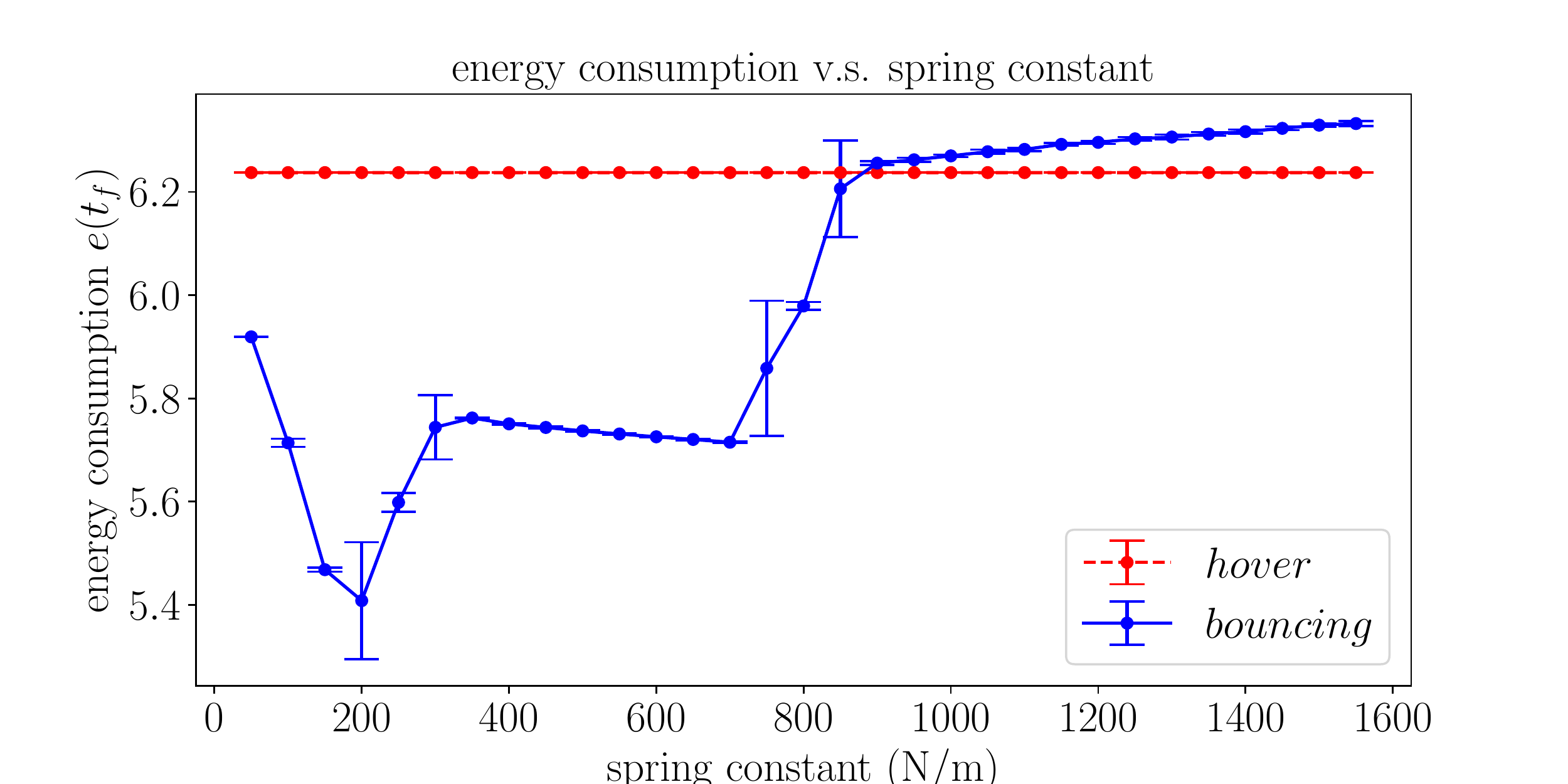}
    \caption{Energy consumption v.s. spring constant. The higher spring constant, the more energy can be stored in the pogo and the stiffer the pogo is.}
    \label{fig:simsc}
\end{figure}
}

\noindent
\paragraph{Damping factor}{
In Section \ref{sec:Model}, we model the pogo as an imperfect spring, which induces energy loss from mechanical motion. The loss is in form of a damping term, which prevents the pogo from being compressed or rebounding quickly and therefore limits the maximum velocity of the PogoDrone after the rebound. 
Fig. \ref{fig:simdf} shows that as the damping factor increases, the energy consumption of the bouncing mode first decreases, then increases and stabilizes at a value lower than that of hovering. We notice that with a perfect spring whose damping factor is ignored or very small, the energy consumption becomes even higher than that of hovering. In theory, when a perfect spring is inserted in the pogo, to recover the same height before falling, the propellers do not need to actuate, since the elastic potential energy stored in the spring is equal to the gravitational energy before falling. However, as described in Section \ref{sec:Control}, the BHC activates the propellers to drive the drone to the desired height $z^d$ upon entering rebound phase despite the state of the pogo. Thus, the redundant energy, provided by the propellers, creates an overshoot in the height, requiring more energy to compensate. 

\begin{figure}[t!]
    \includegraphics[width=8.6cm]{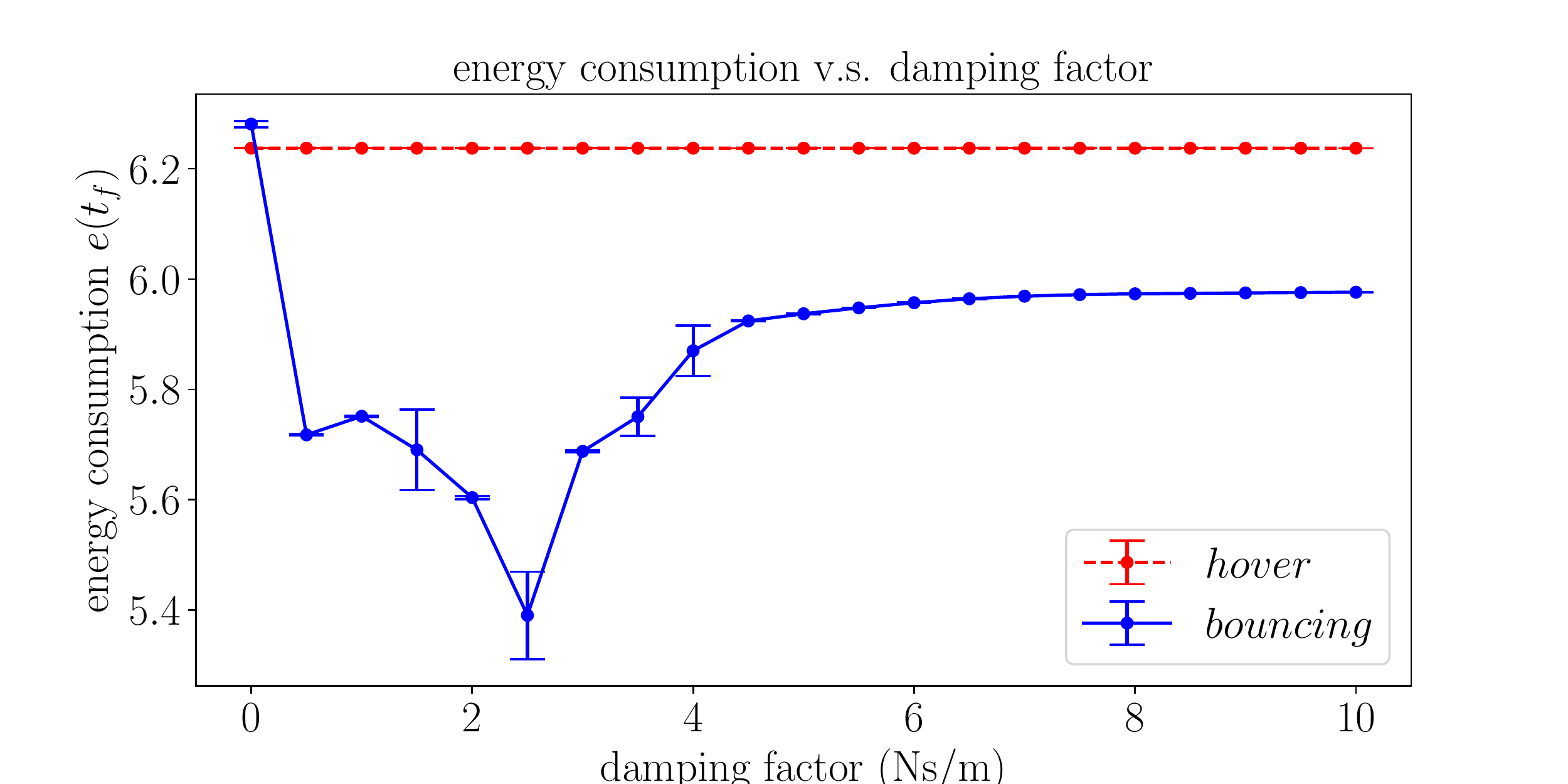}
    \caption{Energy consumption v.s. damping factor.}
    \label{fig:simdf}
\end{figure}
}

\noindent
\paragraph{Desired height}{
The height of the desired position~$\boldsymbol{r}^d$ determines the maximum amount of energy that needs to be stored in the pogo during a bounce. 
As shown in Fig. \ref{fig:simdh}, when the desired height is below 1.5 m, the energy consumption of the PogoDrone is high because of the overshoots created by the controller. As the desired height goes higher, the energy consumption goes down to the low point at 1.5 m when the spring reaches maximum compression when storing all gravitational potential energy. The sudden change between 1.5 m and 2 m is caused by the energy loss due to hard impacts. After 2 m, the trend acts similar to Fig. \ref{fig:simsc}.
{Our simulations verify that energy efficiency is strongly dependent on the spring constant, noise level, and damping coefficient. Fine tuning of the system achieves an optimal amount of energy saved but the jumps should perform under very specific conditions.
}

\begin{figure}[t!]
    \includegraphics[width=8.6cm]{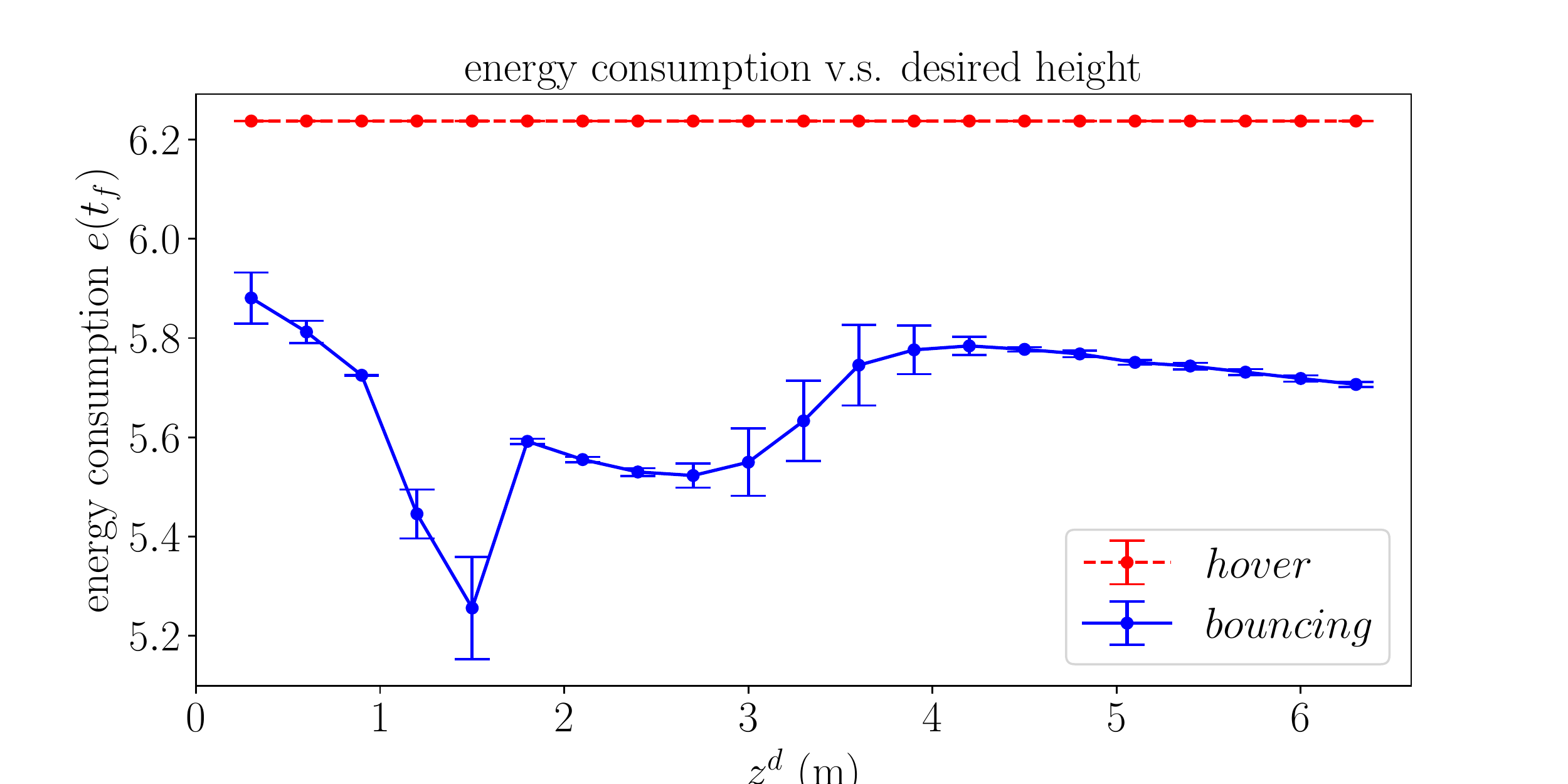}
    \caption{Energy consumption v.s. desired height. The initial desired height determines the initial potential energy.}
    \label{fig:simdh}
\end{figure}
}

}

\subsection{Experiments with an actual robot}

We studied the physical system's efficiency and dynamic performance through five alterations of the desired height.
In our testbed, we used our robot design, described in Sec.~II, and the Crazyflie-ROS package to control the robot's actions. This package communicates with our Optitrack motion capture system, operating at 60 Hz. Our computer runs off-board position and attitude controls through ROS nodes, and these commands are then received by the PogoDrone via a 2.4~GHz radio.
\label{sec:Experiments}
\begin{figure}[t!]
    \includegraphics[width=8.6cm]{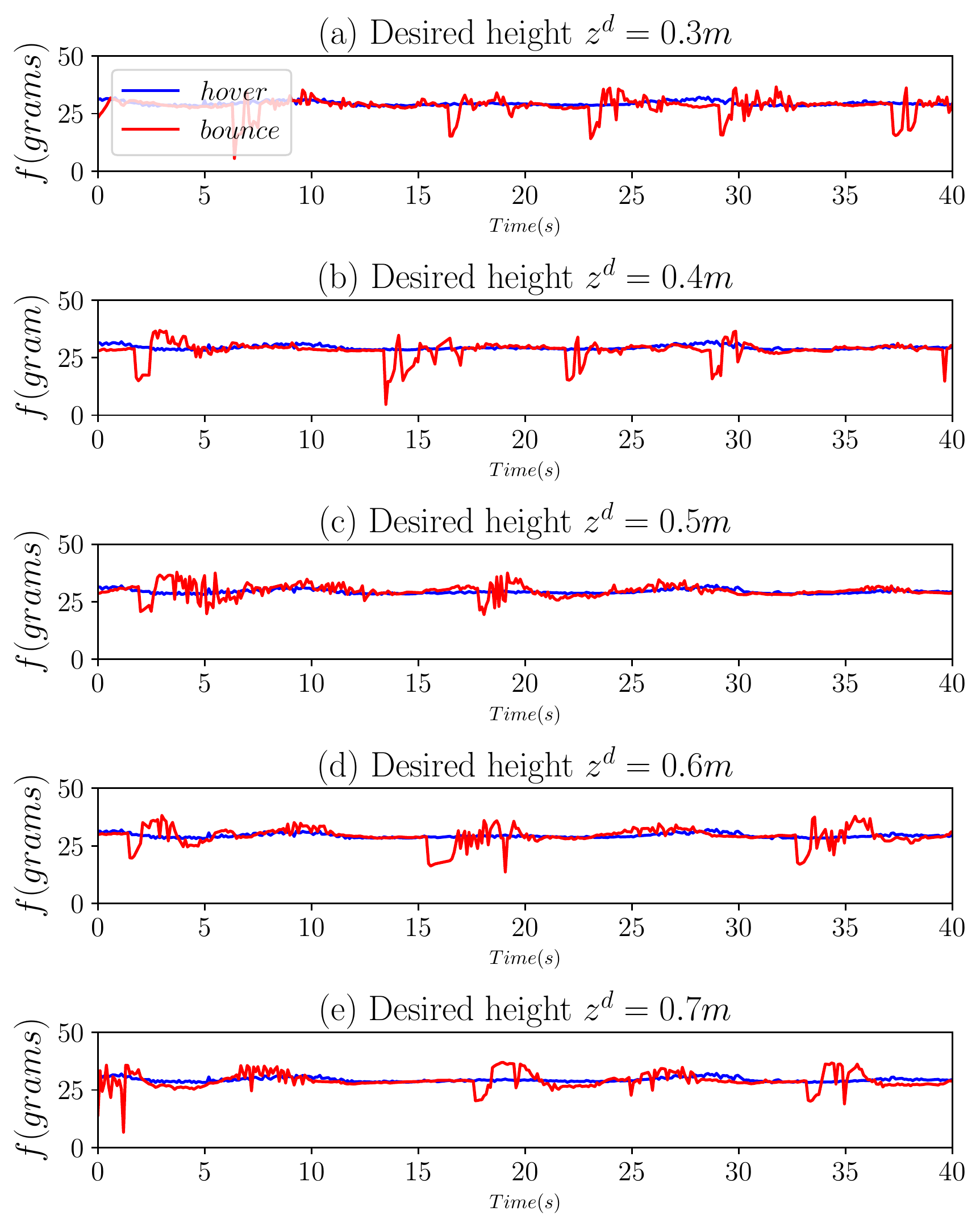}
    \caption{Motor power of the system while hovering and bouncing at each tested height. The top graph shows 0.3 m and each graph below represents the height incremented by 0.1 m.}
    \label{fig:motorpower}
\end{figure}
Fig. \ref{fig:motorpower} shows the force of the motors over time for each height. We find that the optimal height for this bouncing system is at 0.3 m. At this height, energy is not lost to over-compression and the time of recovery in between the bouncing cycle is minimal. As we increase the height, we find that our spring is not able to store the entirety of the potential energy which leads to over-compression. Similar to the simulations, the hard impact with the ground requires the motors to compensate more due to the energy loss. Additionally, the system may generate a torque along the $x$-axis or $y$-axis. This leads to an imperfect bounce which further extends recovery time. At our inflection point of 0.4 m, we find that the culmination of these negative forces reward us with only a slight advantage in efficiency while bouncing.
At 0.5 m, we find that bouncing performs worse than simply hovering. This is primarily due to over-compression, imperfect bounces, and significant recovery time. 
%
At 0.6 and 0.7 m, the efficiency performance is approximately equal for hovering and bouncing. This is caused by an extremely high time of recovery in-between bounces. Over the same time cycle, the PogoDrone spends most of the time recovering which leads to comparable energy performance. Similar to the previous experiments, these high recovery times are caused by over-compression and imperfect bounces along the z-axis. In addition, as the PogoDrone falls from higher heights, our attitude controller must compensate for more air resistance that wishes to create a torque on our system. 
We assumed a perfectly smooth and level surface to bounce on in our experiments. Our surface additionally is rigid which decreases the amount of energy loss per bounce due to damping or absorption.

\section{Conclusion and Future work}
\label{sec:conclusion}
This paper introduces PogoDrone, a  robot that
is able to jump and fly.
We designed a prototype that works in actual environments. 
Using the modeled dynamics, we simulated and analyzed the behavior of the robot after changing different factors.
{We verified the concept of a dual locomotion quadrotor. Our robot is able to rapidly impact with the ground plane and return to a hover state, making it uniquely equipped to handle operations such as taking soil samples or planting seeds. Other multi-rotor vehicles would have to operate close to the ground for a prolonged period of time, making them susceptible to the ground effect.
}
We found that the robot can quickly become energy inefficient
when the spring constant or the actuators' noise is high. 
In conclusion, the robot has the potential to be considerably more efficient than a hovering robot (>20\%), but it requires finding an optimal hardware setup and control gains for the actual robot.


Our robot has the potential to efficiently operate in low heights, so our future work is focused on examining the system's performance on imperfect surfaces which should be explored for real-world applications.


\bibliographystyle{IEEEtran}
\bibliography{ref}

\end{document}